\documentclass[10pt,twocolumn,letterpaper]{article}

\usepackage{cvpr}              

\usepackage{graphicx, amsmath, amssymb, booktabs, caption, subcaption, multirow, overpic, textpos}

\usepackage[pagebackref,breaklinks,colorlinks]{hyperref}

\usepackage[table]{xcolor}

\usepackage{algorithm}
\usepackage{algorithmic}
\usepackage{listings}
\usepackage{bm}

\def\cvprPaperID{****} 
\def\confName{CVPR}
\def\confYear{2023}

\def\eg{\emph{e.g.}} 
\def\ie{\emph{i.e.}}

\def\wrt{w.r.t.}
\def\vs{\emph{vs.}}

\newcommand\ourmethod{RILS}
\newcommand{\maeclip}{MAE+CLIP}

\newlength\savewidth
\newcommand{\tablestyle}[2]{\setlength{\tabcolsep}{#1}\renewcommand{\arraystretch}{#2}\centering\footnotesize}
\renewcommand{\paragraph}[1]{\vspace{1.25mm}\noindent\textbf{#1}}

\newcolumntype{x}[1]{>{\centering\arraybackslash}p{#1pt}}
\newcolumntype{y}[1]{>{\raggedright\arraybackslash}p{#1pt}}
\newcolumntype{z}[1]{>{\raggedleft\arraybackslash}p{#1pt}}

\newcommand{\app}{\raise.17ex\hbox{$\scriptstyle\sim$}}

\definecolor{deemph}{gray}{0.6}

\definecolor{defaultcolor}{gray}{.9}

\begin{document}

\title{RILS: Masked Visual Reconstruction in Language Semantic Space}

\author{Shusheng Yang$^{1,2}$\thanks{This work was done while Shusheng Yang was an intern at ARC Lab, Tencent PCG.}, \ \  Yixiao Ge$^{2}$, \ \  Kun Yi$^{2}$, \ \ Dian Li$^{3}$, \ \ Ying Shan$^{2}$, \ \ Xiaohu Qie$^{4}$, \ \ Xinggang Wang$^{1}$\thanks{Corresponding author, E-mail: {\tt xgwang@hust.edu.cn}.} \\
\normalsize 
$^1$School of EIC, Huazhong University of Science \& Technology \\
\normalsize 
$^2$ARC Lab, $^4$Tencent PCG \qquad $^3$Foundation Technology Center, $^4$Tencent PCG \\
}

\maketitle

\begin{abstract}
Both masked image modeling (MIM) and natural language supervision have facilitated the progress of transferable visual pre-training. In this work, we seek the synergy between two paradigms and study the emerging properties when MIM meets natural language supervision. To this end, we present a novel masked visual Reconstruction In Language semantic Space (\ourmethod{}) pre-training framework, in which sentence representations, encoded by the text encoder, serve as prototypes to transform the vision-only signals into patch-sentence probabilities as semantically meaningful MIM reconstruction targets. The vision models can therefore capture useful components with structured information by predicting proper semantic of masked tokens. Better visual representations could, in turn, improve the text encoder via the image-text alignment objective, which is essential for the effective MIM target transformation. Extensive experimental results demonstrate that our method not only enjoys the best of previous MIM and CLIP but also achieves further improvements on various tasks due to their mutual benefits. \ourmethod{} exhibits advanced transferability on downstream classification, detection, and segmentation, especially for low-shot regimes. Code will be made available at \url{https://github.com/hustvl/RILS}.
\end{abstract}

\section{Introduction}
\label{sec:intro}

Learning transferable representation lies a crucial task in deep learning.
Over the past few years, natural language processing (NLP) has achieved great success in this line of research~\cite{Transformer, ELMo, BERT}.
To explore similar trajectories in the vision domain, researchers tend to draw upon the successes of NLP and have made tremendous progress:
\begin{itemize}
	\item Inspired by the advanced model architecture~\cite{Transformer} as well as self-supervised learning paradigm~\cite{BERT} in NLP, vision Transformers (ViT)~\cite{ViT, Swin} along with masked image modeling (MIM)~\cite{BEiT,MAE} open a new era of self-supervised visual representation learning, and shows attractive transferability on various tasks, especially on fine-grained tasks such as object detection and instance segmentation \cite{ViTDet, MIMDet}.
	\item Inspired by the great scalability brought by leveraging web-scale collections of texts as training data in NLP~\cite{GPT,GPT-2,GPT-3,OPT}, CLIP~\cite{CLIP} and ALIGN~\cite{ALIGN} bring such a principle to vision and manifest the immense potential of leveraging natural language supervision for scalable visual pre-training. Strong transferability of pre-trained visual models on low-shot regimes ensues. It also facilitates diverse application by extracting contextualized image or text features~\cite{ViLD,DALL-E2,DenseCLIP}.
\end{itemize}

\begin{figure}[!t]
\centering
\includegraphics[width=1.0\columnwidth]{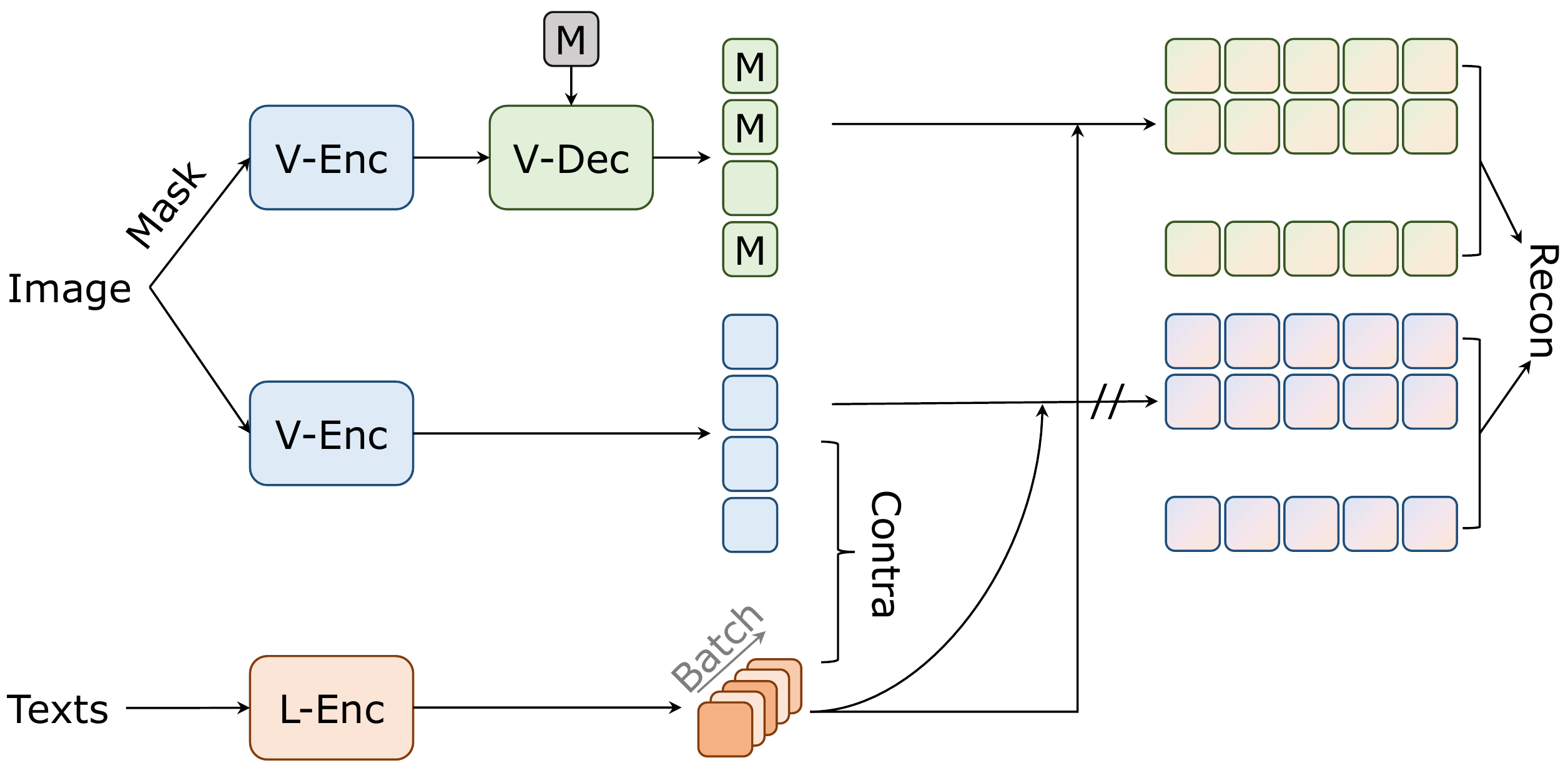}
\caption{\textbf{Overview of our \ourmethod{}.} $\mathrm{Recon}$ and $\mathrm{Contra}$ represent masked reconstruction loss and image-text contrastive loss.
During pre-training, \ourmethod{} learns to perform masked image modeling and image-text contrastive simultaneously.
Masked predictions and corresponding targets are transformed to probabilistic distributions in language space by leveraging text features as semantic-rich prototypes.
Under such a scheme, both objective are unified and achieve mutual benefits from each other.
Vision encoder obtains the ability to capture meaningful and fine-grained context information, sparking decent transfer capacity.
}
\label{fig: architecture}
\vspace{-0.5cm}
\end{figure}

The remarkable attainment achieved by these two lines of research pushes us to ponder: \textit{Is it possible to unify both masked image modeling and natural language supervision to pursuit better visual pre-training?}
A straightforward way towards this goal is to simply combine masked image modeling (MIM) with image-text contrastive learning (ITC) for multi-task learning.
Although the na\"ive combination is feasible to inherit the above advantages, we find it still unsatisfactory since the mutual benefits between MIM and ITC have not yet been fully explored.
Motivated by this, we develop \ourmethod{}, a tailored framework to seek the synergy of masked image modeling and language supervision.

The core insight of \ourmethod{} is to perform \textit{masked visual reconstruction in language semantic space}.
Specifically, instead of reconstructing masked patches in the standalone vision space (\eg, raw pixel~\cite{MAE,SimMIM}, low-level features~\cite{BEiT,MaskFeat} or high-level perceptions~\cite{iBOT,SdAE,MVP,MILAN}), we map patch feature to a probabilistic distribution over a batch of text features as the reconstruction target, which is enabled by ITC that progressively aligns the image and text spaces.
The text features serve as semantically rich prototypes and probabilistic distributions explicitly inject the semantic information onto each image patch.
The MIM objective is formulated as a soft cross-entropy loss to minimize the KL divergence between text-injected probabilistic distributions of masked vision tokens and their corresponding targets.
The visual model optimized by our language-assisted reconstruction objective, in turn, improves ITC with better visual representations that capture fine-grained local contexts.

Under such a working mechanism, the two objectives of MIM and ITC complement each other and form a unified solution for transferable and scalable visual pre-training.
Note that a lot of works \cite{iBOT,MVP,MILAN} have manifested the importance of semantic information in the reconstruction target of MIM objectives.
However, it is abstract to pursue such a semantically rich space with visual-only signals due to its unstructured characteristics \cite{MoCo}. 
Thanks to natural language supervision, this issue is resolved by performing masked reconstruction in language space in our approach.

Extensive experiments on various downstream tasks demonstrate that our design enjoys the best of both worlds.
With a vanilla ViT-B/$16$ as the vision model and $25$-epoch pre-training on $20$ million image-text pairs, \ourmethod{} achieves $83.3\%$ top-$1$ accuracy when fine-tune on ImageNet-$1$K~\cite{ImageNet} classification, $+1.2\%$ and $+0.6\%$ better than the MAE~\cite{MAE} and CLIP~\cite{CLIP} counterparts.
When transferring pre-trained models of \ourmethod{} to fine-grained  tasks such as object detection, instance segmentation, and semantic segmentation, advanced performance can be consistently acquired.
Moreover, our approach exhibits promising results under an extremely low-shot regime.
With only $10$ images per class for linear classification, \ourmethod{} attains $51.8\%$ accuracy on ImageNet-$1$K, indicating strong out-of-the-box capability of our pre-trained model and its generalization ability to real-world scenarios with insufficient annotated data.
\ourmethod{} also demonstrates superior performance on zero-shot image classification and image-text retrieval.
On ImageNet-$1$K benchmark, \ourmethod{} obtains $45.0\%$ zero-shot classification accuracy, $+4.7\%/+3.4\%$ higher than CLIP~\cite{CLIP}$/$SLIP~\cite{SLIP} using the same training data.
Compelling results of \ourmethod{} imply the promising out-of-the-box capacity in the unification of MIM and language supervision.

\section{Related Works}

\noindent\textbf{Masked Image Modeling} translates masked language modeling \cite{BERT} to vision domain and learns transferable visual representation by reconstructing masked signals from partial observation~\cite{BEiT, ViT, iGPT}.
Despite following the same \textit{mask-then-reconstruction} principle, MIM differs from MLM a lot in the design of reconstruction target.
BEiT~\cite{BEiT} utilizes a pre-trained d-VAE~\cite{dVAE,DALL-E} and reconstructs masked image patches in the offline token space.
Subsequent works improve it by employing better pre-trained tokenizer~\cite{PeCo, BEiT2, MVP, MILAN}, eased and refined multi-choice tokens~\cite{mc-BEiT} or contextual aligner~\cite{CAE}.
MAE~\cite{MAE} and SimMIM~\cite{SimMIM} demonstrate directly reconstruct masked patches in raw pixel space can also lead to favorable transferability as well as scalability.
MaskFeat~\cite{MaskFeat} takes hand-crafted low-level HOG feature~\cite{HOG} as target.
Other works like iBOT~\cite{iBOT}, data2vec~\cite{data2vec} and SdAE~\cite{SdAE} perform reconstruction in a high-level vision feature space.
Different from these methods, in this work, we tap the potential when masked image modeling meets natural language supervision and propose performing masked visual reconstruction in the language semantic space.

\noindent\textbf{Language Supervised Visual Pre-training} learns visual representation from image-text pairs by solving generative \cite{ICMLM, VirTex} or discriminative \cite{ConVIRT} pretext tasks.
Recently, benefit from modern scalable networks \cite{ViT, Swin, Swinv2} and public available image-text datasets \cite{CC3M, CC12M, RedCaps, LAION-400M, LAION-5B}, CLIP~\cite{CLIP} and ALIGN~\cite{ALIGN} unveil the tremendous transferability and scalability of this paradigm.
The core technique of CLIP is aligning both vision and language modalities in a joint embedding space by global representation contrastive.
Follow-up works further improve CLIP from the vision-only \cite{UniCL, SLIP} or vision-language \cite{FILIP, DeCLIP, OTTER, MSCLIP} side.
In this paper, we bring natural language supervision together with masked image modeling for better visual pre-training on these two paradigms.

\section{Our Approach}

\subsection{Architecture}

Among numerous architecture designing spaces, without loss of generalization, we adopt an asymmetric \textit{encoder-decoder} architecture following MAE~\cite{MAE} and a \textit{dual-encoder} architecture following CLIP~\cite{CLIP} for their flexibility.
As illustrated in Figure~\ref{fig: architecture}, \ourmethod{} comprises the following three major components:

\noindent\textbf{Vision Encoder}
plays the key role in \ourmethod{} and all our strive aims to strengthen its capacity on downstream transfer.
Following the trend in recent visual pre-training, we implement this encoder by a vanilla vision Transformer (ViT)~\cite{ViT}.
It takes both original (unmasked) image and corrupted (masked) image as inputs.
Formally, input image $I$ is first divided into regular non-overlapping image patches and then encoded by a stack of Transformer blocks~\cite{Transformer}.
Meanwhile, following MAE~\cite{MAE}, we randomly mask a large portion of image patches and leave the remaining patches to be visible.
This corrupted image $\hat{{I}}$ is also encoded by vision encoder.
We formulate the process of vision encoder as:
\begin{equation}
\begin{aligned}
\label{equation: vision_encoder}
\mathrm{V\text{-}Enc}({I}) &= \{{f}^{{k}}|{k}\in[1,N]\},\\
\mathrm{V\text{-}Enc}(\hat{{I}}) &= \{\hat{{f}}^{{k}}|{k}\in [1,N]\backslash \mathcal{M}\},
\end{aligned}
\end{equation}
in which $k$ denotes the patch index and $N$ denotes image patch numbers.
$f$ and $\hat{f}$ betoken encoded patch features of original image $I$ and masked image $\hat{I}$.
$\mathcal{M}$ indicates the index set of random masked patches.

\noindent\textbf{Language Encoder}
encodes input text $T$ by a stack of Transformer layers with causal masked attention \cite{Transformer}.
This process can be simply represented by:
\begin{equation}
\label{equation: text_encoder}
\mathrm{L\text{-}Enc}({T}) = {h}.
\end{equation}
We take the output $h$ as global representation of input text.

\noindent\textbf{Vision Decoder} consists of another series of Transformer blocks.
Particularly, in our design, decoder blocks have the same dimension as encoder blocks.
It takes the encoded feature $\hat{f}$ of masked image $\hat{I}$ along with a learnable $\mathrm{[MASK]}$ token as inputs, and tries to predict masked signals from corrupted view:
\begin{equation}
\label{equation: vision_decoder}
\mathrm{V\text{-}Dec}(\{\hat{{f}}^{{k}}|{k}\in [1,N]\backslash \mathcal{M}\}, \mathrm{[MASK]}) = \{{g}^{{k}}|{k}\in[1,N]\}.
\end{equation}

\subsection{Training Objective}

\noindent\textbf{Image-Text Contrastive.}
We leverage image-text contrastive loss to align both modality in a joint embedding space.
Specifically, given image-text pair $\{({I}, {T})\}$, we take the mean-pooled image feature $\bar{f}=\frac{1}{N}\sum_{k=1}^{N}{f^{k}}$ and ${h}$ in Eq.~\eqref{equation: text_encoder} as global representation for image and text.
The image and text features are further projected by two projection heads and followed by a normalization:
\begin{equation}
\label{equation: normalize_feature}
\begin{aligned}
	{z}^{I} &= ||\theta(\bar{f})||, \\
	{z}^{T} &= ||\phi(h)||,
\end{aligned}
\end{equation}
$\theta(\cdot)$ and $\phi(\cdot)$ denotes the projection head for image and text respectively.
The image-to-text contrastive loss and text-to-image contrastive loss can be represented as:
\begin{equation}
\begin{aligned}
\mathcal{L}_\mathrm{I2T} &= -\frac{1}{B}\sum_{i=1}^B\log\frac{\exp(\langle z^I_i, z^T_i \rangle/ \sigma)}{\sum_{j=1}^B \exp(\langle z^I_i, z^T_j \rangle/ \sigma)}, \\
\mathcal{L}_\mathrm{T2I} &= -\frac{1}{B}\sum_{i=1}^B\log\frac{\exp(\langle z^T_i, z^I_i \rangle/ \sigma)}{\sum_{j=1}^B \exp(\langle z^T_i, z^I_j \rangle/ \sigma)},
\end{aligned}
\end{equation}
in which $i$ and $j$ stands for the index within a mini-batch and $B$ indicates the batch size respectively.
$\sigma$ performs a learnable temperature and is jointly trained during the pre-training.
The total loss of image-text contrastive learning can be formulated as:
\begin{equation}
\label{equation: total_loss}
	\mathcal{L}_{\mathrm{Contra}} = \frac{1}{2}(\mathcal{L}_\mathrm{I2T} + \mathcal{L}_\mathrm{T2I}).
\end{equation}

\noindent\textbf{Masked Visual Reconstruction in Language Semantic Space.}
As aforementioned, despite the \textit{mask-then-reconstruct} principle of MIM is concise enough, the contiguous and unstructured characteristics in visual signal make the choice of reconstruction space non-trivial.
Lots of works have manifested the great importance of performing masked reconstruction in a semantic-rich space~\cite{iBOT, SdAE, MVP, MILAN}.
In this work, we build our reconstruction space from a vision-language perspective.
We regard the text features as natural semantic descriptors for image patches and try to perform masked visual reconstruction in this language space.
Specifically, given the encoded patch feature ${f}^{k}$ in Eq.~\eqref{equation: vision_encoder} with $k$ being the index of patch and decoded patch feature ${g}^{k}$ in Eq.~\eqref{equation: vision_decoder}, we firstly project and normalize both features to the vision-language aligned space:
\begin{equation}
\begin{aligned}
	\tilde{f}^{k}_{i} &= ||\theta{({f}^{k}_{i})}||, \\
	\tilde{g}^{k}_{i} &= ||\theta{({g}^{k}_{i})}||,
\end{aligned}
\end{equation}
with $i$ being the index within mini-batch. $\theta(\cdot)$ represents the same vision projection head in Eq.~\eqref{equation: normalize_feature}.
The key step of our design is to map patch features to a probabilistic distributions over a bunch of text features:
\begin{equation}
\label{equation: prob_distribution}
\begin{aligned}
	\bm{p}^{k}_{i} &= \{\frac{\exp(\langle \tilde{f}^{k}_{i}, {z}_{l}^{T}\rangle/\tau_{1})}{\sum_{j=1}^{B}{\exp(\langle \tilde{f}^{k}_{i}, {z}_{j}^{T}\rangle/\tau_{1})}} \ | \ l \in[1,B]\}, \\
	\bm{q}^{k}_{i} &= \{\frac{\exp(\langle \tilde{g}^{k}_{i}, {z}_{l}^{T}\rangle/\tau_{2})}{\sum_{j=1}^{B}{\exp(\langle \tilde{g}^{k}_{i}, {z}_{j}^{T}\rangle/\tau_{2})}} \ | \ l \in[1,B]\},
\end{aligned}
\end{equation}
in which $\tau_{1}$ and $\tau_{2}$ serve as temperatures.
In this way, with the text features serve as semantic-rich prototypes, both masked prediction and corresponding target are mapped into this language semantic space.
The probabilistic distributions explicitly express the context information within each patch.
The reconstruction objective is to shrinking the differences between text-injected distributions of target and masked prediction by minimize the KL divergence of $p_{i}^{k}$ \wrt{} $q_{i}^{k}$, which can be represented by:
\begin{equation}
\begin{aligned}
	\mathcal{L}_\mathrm{Recon} = \frac{1}{||\mathcal{C}||\cdot||\mathcal{M}||}\sum_{i\in \mathcal{C}}\sum_{k\in\mathcal{M}}-\mathrm{sg}[\bm{p}^{k}_{i}]\log\bm{q}^{k}_{i},
\end{aligned}
\end{equation}
in which $\mathrm{sg}[\cdot]$ indicates stop gradient.
$\mathcal{M}$ denotes the index set of masked patches.
$\mathcal{C}$ signifies the indexes of images which are correctly aligned to corresponding text features.
In other words, we only calculate reconstruction loss on images which are correctly matched with target texts in image-to-text matching.

By transferring reconstruction space from standalone vision space to language space, our approach takes both MIM and ITC into a unifying landscape and achieves mutual benefits from each other.
MIM always suffers from overly paying attention on low-level details which consume lots of model's capacity but of less helpful for understanding visual concepts.
By leveraging text features as prototypes and transfer patch features to probabilistic distribution on language space, the low-level information inside visual signals are abandoned by the contextualized language prototypes.
Better contextualized image features in turn assist vision-language alignment, leading to transferable and scalable visual pre-training.

\noindent\textbf{Overall Objective Function.}
The final objective of \ourmethod{} is a weighted-sum of both image-text contrastive loss and masked reconstruction loss:
\begin{equation}
	\mathcal{L}_\mathrm{\ourmethod} = \lambda_{1}\cdot\mathcal{L}_\mathrm{Contra} + \lambda_{2}\cdot\mathcal{L}_\mathrm{Recon}.
\end{equation}
$\lambda_{1}$ and $\lambda_{2}$ indicate coefficients to balance two losses.

\begin{figure}[!t]
\centering
\includegraphics[width=1.0\columnwidth]{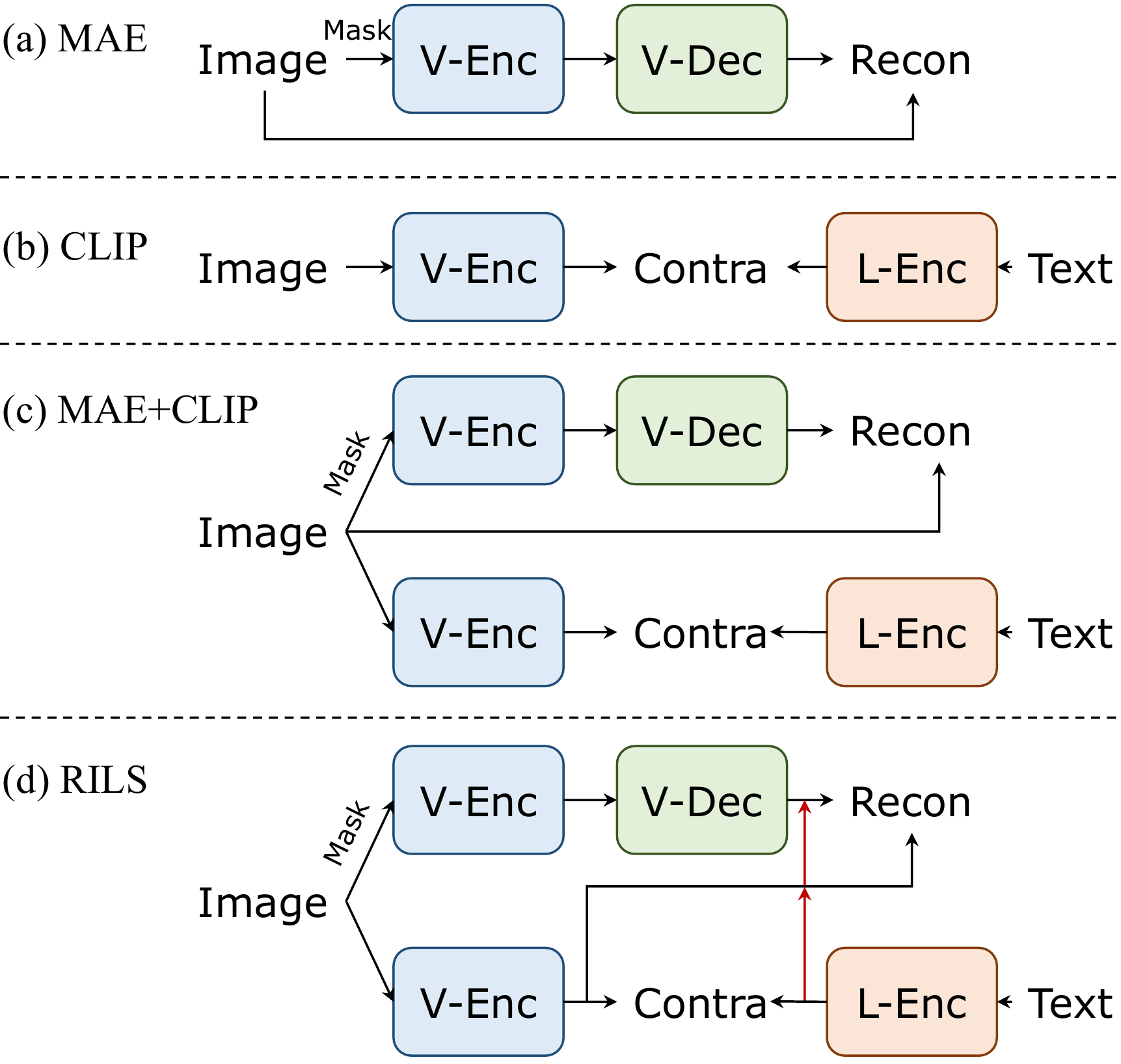}
\caption{Architecture comparisons between MAE~\cite{MAE}, CLIP~\cite{CLIP}, \maeclip{} and \ourmethod{}. $\mathrm{Recon}$ and $\mathrm{Contra}$ indicates masked reconstruction loss and image-text contrastive loss.}
\label{fig: comparison}
\vspace{-0.5cm}
\end{figure}

\subsection{Pre-training Setup}

Similar to \cite{MSCLIP}, we sequentially sample image-text pairs according to filenames from recent released LAION-$400$M~\cite{LAION-400M} dataset as our training sets.
We term them as L-$10$M$/$L-$20$M$/$L-$50$M according to the amount of sampled unique image-text pairs (\eg, L-$10$M stands for the first $10$ million subset of LAION-$400$M).
Unless specified, our method is trained on L-$20$M for $25$ epochs.
We take AdamW~\cite{Adam} as optimizer with learning rate set to $5$e-$4$ and a weight decay of $0.5$.
Learning rate linearly increases in the first epoch as warmup and decreases in the rest following the cosine learning rate decay strategy.
We train our method on $32$ NVIDIA V$100$ with a total batch size of $4096$ (\ie, batch size per GPU is $128$).
For model architecture, we take the widely-adopted ViT-B/$16$ as vision encoder, $1$-layer Transformer block with $768$-dim and $12$ heads as vision decoder, and a text Transformer with $12$ blocks and $512$-dim as language encoder.
To tokenize text inputs, following \cite{CLIP}, we use byte-pair encoding (BPE~\cite{BPE}) with $49$K vocabulary size and set the max length of each sentence to $77$.
During pre-training, input images are resized to $224\times224$ and we set random mask ratio to $75\%$ following \cite{MAE}.
Temperatures $\tau_{1}$ and $\tau_{2}$ in Eq.~\eqref{equation: prob_distribution} are set to $0.04$ and $0.1$.
Loss coefficients $\lambda_{1}$ and $\lambda_{2}$ are set to $1.0$ and $0.5$ by default.
More pre-training setups are listed in the appendix.

\subsection{Discussion}

There are also meaningful attempts~\cite{MVP, MILAN, BEiT2, MaskCLIP} on utilizing natural language supervision together with MIM and seem alike to ours.
However, there still have some distinctions existing in the motivation and method between ours and theirs.
MVP~\cite{MVP}, MILAN~\cite{MILAN} and BEiTv$2$~\cite{BEiT2} leverage natural language supervision by a two-stage framework, while ours is fully end-to-end.
We will take further discussion about two-stage methods and ours in later experiments.
A related and concurrent work \cite{MaskCLIP} shows some similar designs but significantly differs from our idea of reconstruction in language space. As it does not have a reproducible implementation, we do not take it into consideration for comparison.

\section{Main Results}

\begin{table}
\centering
\tablestyle{3pt}{1.1}
\setlength{\tabcolsep}{0.9mm}{
\begin{tabular}{l|cc|c|c}
Method & Dataset & PT Epo. & Lin. & FT. \\
\toprule
SimCLR~\cite{SimCLR} & \multirow{6}{*}{L-$20$M} & \multirow{6}{*}{$25$($\sim400$)} & $51.7$ & $81.3$ \\
MAE~\cite{MAE} & & & $44.3$ & $82.1$ \\
CLIP~\cite{CLIP} & & & $67.8$ & $82.7$ \\
SLIP~\cite{SLIP} & & & $70.1$ & $82.6$ \\
\maeclip{} & & & $64.5$ & $82.9$ \\
\ourmethod{} & & & $\mathbf{\underline{71.5}}$ & $\mathbf{\underline{83.3}}$ \\
\midrule
BEiT~\cite{BEiT} & \multirow{2}{*}{IN-$1$K($\sim1.3$M)} & $800$ & $-$ & $83.2$ \\
MAE~\cite{MAE} & & $1600$ & $67.8$ & $83.6$ \\
\ourmethod{} & L-$50$M & $25$($\sim1000$) & $\mathbf{\underline{71.9}}$ & $\mathbf{\underline{83.6}}$ \\
\bottomrule
\end{tabular}}
\caption{\textbf{Image classification results on ImageNet-$\mathbf{1}$K (IN-$\mathbf{1}$K).} PT Epo. indicates per-training epochs. Lin. and FT. is short for linear probing and end-to-end fine-tuning respectively.}
\label{tab: imagenet_classification}
\vspace{-0.5cm}
\end{table}

\begin{table*}[tbp]
\centering
\tablestyle{3pt}{1.1}
\setlength{\tabcolsep}{2.5mm}{
\begin{tabular}{l|ccc|ccc|ccc|ccc}
\multirow{2}{*}{Method} & \multicolumn{6}{c}{COCO} & \multicolumn{6}{c}{LVIS} \\
\cmidrule(lr){2-7}\cmidrule(lr){8-13}
& AP$^{\mathcal{B}}$ & AP$^{\mathcal{B}}_{50}$ & AP$^{\mathcal{B}}_{75}$ & AP$^{\mathcal{M}}$ & AP$^{\mathcal{M}}_{50}$ & AP$^{\mathcal{M}}_{75}$ & AP$^{\mathcal{B}}$ & AP$^{\mathcal{B}}_{50}$ & AP$^{\mathcal{B}}_{75}$ & AP$^{\mathcal{M}}$ & AP$^{\mathcal{M}}_{50}$ & AP$^{\mathcal{M}}_{75}$ \\
\toprule
MAE~\cite{MAE} & $48.1$ & $68.6$ & $52.9$ & $42.4$ & $65.8$ & $\mathbf{\underline{46.4}}$ & $31.0$ & $46.2$ & $33.7$ & $29.6$ & $44.0$ & $31.7$ \\
CLIP~\cite{CLIP} & $47.7$ & $69.1$ & $52.3$ & $42.0$ & $65.9$ & $45.1$ & $32.3$ & $48.1$ & $35.1$ & $30.5$ & $45.9$ & $32.5$ \\
SLIP~\cite{SLIP} & $46.5$ & $68.5$ & $51.0$ & $41.5$ & $65.1$ & $44.1$ & $32.4$ & $48.9$ & $35.1$ & $30.8$ & $46.5$ & $32.6$ \\
\maeclip{} & $48.1$ & $69.6$ & $52.5$ & $42.4$ & $66.2$ & $45.7$ & $32.6$ & $48.8$ & $35.2$ & $30.7$ & $46.4$ & $32.6$ \\
\midrule
\ourmethod & $\mathbf{\underline{48.5}}$ & $\mathbf{\underline{70.5}}$ & $\mathbf{\underline{53.2}}$ & $\mathbf{\underline{42.6}}$ & $\mathbf{\underline{66.8}}$ & $45.8$ & $\mathbf{\underline{33.8}}$ & $\mathbf{\underline{50.2}}$ & $\mathbf{\underline{36.5}}$ & $\mathbf{\underline{31.6}}$ & $\mathbf{\underline{47.8}}$ & $\mathbf{\underline{33.7}}$ \\
\bottomrule
\end{tabular}}
\vspace{-0.2cm}
\caption{\textbf{Object detection and instance segmentation results on COCO and LVIS.} All models are pre-trained with ViT-B/$16$ for $25$ epochs on L-$20$M. Fine-tuning recipes for different pre-trained models are the same.}
\vspace{-0.5cm}
\label{tab: object_detection}
\end{table*}

In this section, we evaluate the representation quality of pre-training by transferring pre-trained models to various downstream tasks.
We choose MAE~\cite{MAE} and CLIP~\cite{CLIP} as representative methods of masked image modeling and vision-language contrastive learning.
We also conduct a na\"ive baseline (termed as \maeclip{}) which simply combine MAE and CLIP together to perform MIM and ITC simouteneously, as a counterpart for our approach.
Semantically comparisons between MAE, CLIP, \maeclip{} and our \ourmethod{} are illustrated in Figure~\ref{fig: comparison}.

\subsection{Classification Transfer}

\noindent\textbf{Linear Probing} evaluates the quality of pre-trained feature by training a linear classifier on the frozen feature.
We following the the recipe in \cite{DINO} and sweep over different learning rate.
Results are shown in Table~\ref{tab: imagenet_classification}.
We notice that, with $25$-epochs pre-training on L-$20$M, \maeclip{} only achieves $64.5\%$ accuracy which is better than MAE but worse than CLIP.
This implies the contradiction between MIM and ITC exists in such a na\"ive combination.
\ourmethod{} alleviates this contradiction by elaborated design and outperforms other methods by a large margin.

\noindent\textbf{End-to-End Fine-tuning.}
We follow most of setups in \cite{MAE} for fine-tuning.
Concretely, we fine-tune pre-trained models for $100$ epochs with a warm-up of $5$ epochs.
Hyper-parameters are all the same for all experiments except the learning rate.
Results are shown in Table~\ref{tab: imagenet_classification}.
When training with $25$ epochs on L-$20$M, our method shows distinct advantages.
Compared to MAE, CLIP and \maeclip{}, our method exhibits $+1.2\%$, $+0.6\%$ and $+0.4\%$ gains respectively.
Moreover, when we scale-up the dataset capacity from L-$20$M to L-$50$M, with $25$ epochs pre-training (around $1000$ equivalent epochs in the ImageNet-$1$K regime), our method achieves $83.6\%$ top-$1$ accuracy which is on par with prior art (MAE trained on ImageNet-$1$K with $1600$ epochs) with only $62.5\%$ training length.

\subsection{Downstream Transfer}

\noindent\textbf{Object Detection and Segmentation.}
For object detection and instance segmentation, we choose COCO~\cite{COCO} and LVIS~\cite{LVIS} as benchmarks.
We follow the design in \cite{ViTDet} to transfer pre-trained ViT to detection.
To tame quadratic complexity within self-attention, most attention blocks in the ViT are replaced with window attention except for four global blocks to perform cross-window interaction.
SimpleFPN~\cite{ViTDet} is attached to the last transform block to generate pyramid features.
Modernized RPN~\cite{FasterR-CNN} and Mask R-CNN~\cite{MaskR-CNN} head are deployed for detecting and segmenting visual instances.
All pre-trained models are fine-tuned on two benchmarks for $25$ epochs with same hyper-parameters.
The results are shown in Table~\ref{tab: object_detection}.
Among all methods, our \ourmethod{} achieves the best results in terms of AP$^{\mathcal{B}}$ and AP$^{\mathcal{M}}$ on both COCO and LVIS.
It's noteworthy that two benchmarks show different properties: COCO benchmark shows less benefits from language supervision while LVIS converses.
Specifically, MAE shows leading performance on COCO but inferior performance on LVIS.
We suspect this is due to the inherent distinctions in COCO and LVIS: LVIS contains $1203$ visual categories which is about $15\times$ more than COCO, and it always suffers from the long-tail distribution.
Under such circumstances, COCO requires more localization ability which MAE excel at while LVIS prefers better classification ability which natural language supervision can bring.
From this perspective, when compare our \ourmethod{} with the \maeclip{}, we find our design benefits more from both MIM and ITC objectives.
On both COCO and LVIS, \maeclip{} only shows competitive performance to the winner of MAE and CLIP, but our \ourmethod{} exhibits apparent improvements especially on LVIS.
This indicates our design leverage masked image modeling and language supervision in a more synergistic way.
We believe this kind of synergy is of great exploration value for better visual pre-training.

\begin{table}
\centering
\tablestyle{3pt}{1.1}
\setlength{\tabcolsep}{2.5mm}{
\begin{tabular}{l|cc|c}
Method & Dataset & PT Epo. & mIoU \\
\toprule
\multirow{2}{*}{BEiT~\cite{BEiT}} & \multirow{4}{*}{IN-$1$K($\sim1.3$M)} & $300$ & $45.5$ \\
& & $800$ & $46.5$ \\
\multirow{2}{*}{MAE~\cite{MAE}} & & $300$ & $45.8$ \\
& & $1600$ & $48.1$ \\
\midrule
MAE~\cite{MAE} & \multirow{4}{*}{L-$20$M} & \multirow{4}{*}{$25$($\sim400$)} & $44.2$ \\
CLIP~\cite{CLIP} & & & $45.2$ \\
SLIP~\cite{SLIP} & & & $45.7$ \\
\maeclip{} & & & $45.3$ \\
\midrule
\ourmethod{} & \multirow{1}{*}{L-$20$M} & \multirow{1}{*}{$25$($\sim400$)} & $\mathbf{\underline{48.1}}$ \\
\bottomrule
\end{tabular}}
\vspace{-0.2cm}
\caption{\textbf{Semantic segmentation results on ADE$\mathbf{20}$K.}}
\label{tab: ade20k_semantic_segmentation}
\vspace{-0.5cm}
\end{table}

\begin{table*}[tbp]
\centering
\tablestyle{3pt}{1.1}
\setlength{\tabcolsep}{1.5mm}{
\begin{tabular}{l|cc|x{64}x{64}x{64}x{64}}
\multirow{2}{*}{Method} & \multirow{2}{*}{PT Dataset} & \multirow{2}{*}{PT Epo.} & \multicolumn{4}{c}{Images per Class} \\
\cmidrule(lr){4-7} & & & $1$ & $2$ & $5$ & $10$ \\
\toprule
\multirow{1}{*}{MAE}~\cite{MAE} & \multirow{1}{*}{IN-$1$K($\sim1.3$M)} & \multirow{1}{*}{$1600$} & $4.3\pm(0.28)$ & $10.6\pm(0.21)$ & $22.4\pm(0.28)$ & $31.6\pm(0.02)$ \\
\midrule
\multirow{1}{*}{BEiT}~\cite{BEiT} & \multirow{1}{*}{IN-$1$K($\sim1.3$M)} & $800$ & $1.3\pm(0.03)$ & $2.2\pm(0.13)$ & $4.4\pm(0.21)$ & $7.4\pm(0.05)$ \\
\midrule
MAE~\cite{MAE} & \multirow{4}{*}{L-$20$M} & \multirow{4}{*}{$25(\sim400)$} & $3.4\pm(0.12)$ & $5.2\pm(0.21)$ & $10.1\pm(0.10)$ & $14.8\pm(0.20)$ \\
CLIP~\cite{CLIP} & & & $19.4\pm(0.18)$ & $29.2\pm(0.61)$ & $39.8\pm(0.39)$ & $46.3\pm(0.15)$ \\
SLIP~\cite{SLIP} & & & $17.7\pm(0.33)$ & $27.2\pm(0.56)$ & $38.6\pm(0.55)$ & $46.4\pm(0.06)$ \\
\maeclip{} & & & $21.1\pm(0.12)$ & $31.1\pm(0.94)$ & $41.6\pm(0.43)$ & $47.5\pm(0.15)$ \\
\midrule
\multirow{1}{*}{\ourmethod} & \multirow{1}{*}{L-$20$M} & $25(\sim400)$ & $\mathbf{\underline{24.0}}\pm(0.27)$ & $\mathbf{\underline{34.6}}\pm(0.88)$ & $\mathbf{\underline{45.7}}\pm(0.46)$ & $\mathbf{\underline{51.8}}\pm(0.18)$  \\
\bottomrule
\end{tabular}}
\vspace{-0.2cm}
\caption{\textbf{Extreme low-shot classification on ImageNet-$\mathbf{1}$K.} We random sample $1$, $2$, $5$, $10$ images per class from training split, and report logistic regression accuracy ($\%$) on ImageNet-$1$K validation split. All methods use ViT-B/$16$ as vision encoder.}
\label{tab: extreme_low_shot_imagenet}
\vspace{-0.5cm}
\end{table*}

\noindent\textbf{Semantic Segmentation.}
Experiments on semantic segmentation are conducted on the well-known ADE$20$K~\cite{ADE20K} dataset.
We build the segmentation framework upon UperNet~\cite{UPerNet} and use the pre-trained models as encoders.
Input images are resized to $512\times512$ and all models are fine-tuned for $160$K iterations.
All hyper-parameters strictly follow MAE~\cite{MAE} and not tuned.
We report the mean intersection-over-union (mIoU) in Table~\ref{tab: ade20k_semantic_segmentation}.
As the results shown, our method overwhelmingly surpasses others.
Specifically, with $25$ epoch pre-training on L-$20$M, our method achieves $48.1$ mIoU, $+3.9$ and $+2.9$ higher than MAE and CLIP.
Similar to the trends on LVIS, \maeclip{} only gets $0.1$ mIoU gains by simply combining MIM with ITC together, far less than our approach.
Furthermore, our method shows competitive or better performance when compared to prior art.
Compared to MAE pre-trained on ImageNet-$1$K with $300$ epochs, our method achieves $2.3$ higher performance ($48.1$ \vs \ $45.8$).
When MAE is pre-trained with $1600$ epochs, our method achieves the same mIoU ($48.1$) while only requires $25\%$ training length.

Experiments above demonstrate the excellent transfer capacity of our approach on fine-grained visual understanding tasks.
Our design unleashes the ability to capture local details and global contexts by performing masked visual reconstruction in language semantic space.

\subsection{Label-Efficient Transfer}

\noindent\textbf{Low-shot Regime Classification.}
We investigate the classification performance when only very few labeled images are available.
Specifically, following \cite{SwAV, MSN}, we random sample $1$, $2$, $5$, and $10$ labeled images per class from ImageNet-$1$K training split as our training sets.
Instead of end-to-end fine-tuning, we train a linear classifier on frozen features to avoid overfitting.
The complete validation split of ImageNet-$1$K which contains $50$K images are used to evaluate the accuracy.
Table~\ref{tab: extreme_low_shot_imagenet} shows the results.

Specifically, compared to \maeclip{} which only obtain slightly improvements over CLIP, our \ourmethod{} outperforms both of them by a large margin.
Notably, with only $10$ images per class (\ie, $10$K images for $1$K classes), our method can achieve $51.8\%$ top-$1$ accuracy.

\noindent\textbf{Low-shot Regime Detection.}
We further transfer the low-shot experiment to object detection on COCO~\cite{COCO}, which requires model to localize and classify visual objects simultaneously.
We randomly sample annotated images from COCO training split with different sampling ratio (range from $1\%$ to $50\%$) as our training sets.
All models are end-to-end fine-tuned for $12$ epochs instead of $25$ to prevent overfitting.
We report the average precision of detection AP$^{\mathcal{B}}$ for comparison in Table~\ref{tab: extreme_low_shot_detection}.
As the results shown, our method shows the best performance under a wide range of sampling ratio (from $2\%$ to $50\%$).

Conceptually, one of the aspirations of pre-training is to pursue efficient transfer (\eg, less trainable data, shorter training length) on downstream tasks~\cite{BYOL, MSN, he2019rethinking}.
Experiments in the low-shot regime show the strong out-of-the-box capacity of our \ourmethod{} by performing MIM in language semantic space.
The non-trivial results indicate our pre-training approach brings out label-efficient learner, showing great application value to real-world scenarios, especially when annotated data is insufficient.

\begin{table}[tbp]
\centering
\tablestyle{3pt}{1.1}
\setlength{\tabcolsep}{1mm}{
\begin{tabular}{l|x{24}x{24}x{24}x{24}x{24}x{24}}
\multirow{2}{*}{Method} & \multicolumn{6}{c}{COCO Sampling Ratio} \\
\cmidrule(lr){2-7} & $1\%$ & $2\%$ & $5\%$ & $10\%$ & $20\%$ & $50\%$ \\
\toprule
MAE~\cite{MAE} & $0.94$ & $6.10$ & $15.76$ & $23.16$ & $29.78$ & $38.10$ \\
CLIP~\cite{CLIP} & $0.81$ & $5.05$ & $14.98$ & $22.49$ & $29.88$ & $38.50$ \\
SLIP~\cite{SLIP} & $\mathbf{\underline{1.11}}$ & $4.54$ & $13.84$ & $21.91$ & $29.53$ & $37.73$ \\
\maeclip{} & $0.68$ & $5.28$ & $14.33$ & $23.72$ & $29.99$ & $39.24$ \\
\midrule
\multirow{1}{*}{\ourmethod} & $0.86$ & $\mathbf{\underline{6.46}}$ & $\mathbf{\underline{16.94}}$ & $\mathbf{\underline{24.69}}$ & $\mathbf{\underline{31.97}}$ & $\mathbf{\underline{40.41}}$ \\
\bottomrule
\end{tabular}}
\vspace{-0.2cm}
\caption{\textbf{Low-shot regime object detection on COCO}. We report detection performance AP$^\mathcal{B}$ with $12$ epochs fine-tuning. All models are pre-trained with ViT-B/$16$ and $25$ epochs on L-$20$M.}
\vspace{-0.6cm}
\label{tab: extreme_low_shot_detection}
\end{table}

\subsection{Zero-Shot Transfer}

\begin{table*}[!tbp]
\centering
\tablestyle{3pt}{1.0}
\footnotesize
\setlength{\tabcolsep}{0.5mm}{
\begin{tabular}{l|ccccccccccccccccccccc|cc}
Method & \rotatebox{90}{\scriptsize Food$101$} &  \rotatebox{90}{\scriptsize CIFAR$10$} & \rotatebox{90}{\scriptsize CIFAR$100$} & \rotatebox{90}{\scriptsize CUB$200$} & \rotatebox{90}{\scriptsize SUN$397$} & \rotatebox{90}{\scriptsize Cars} & \rotatebox{90}{\scriptsize Aircraft} & \rotatebox{90}{\scriptsize DTD} & \rotatebox{90}{\scriptsize Pets} & \rotatebox{90}{\scriptsize Caltech$101$} & \rotatebox{90}{\scriptsize Flowers} & \rotatebox{90}{\scriptsize MNIST} & \rotatebox{90}{\scriptsize FER$2013$} & \rotatebox{90}{\scriptsize STL$10$} & \rotatebox{90}{\scriptsize EuroSAT} & \rotatebox{90}{\scriptsize RESISC$45$} & \rotatebox{90}{\scriptsize GTSRB} & \rotatebox{90}{\scriptsize Country$211$} & \rotatebox{90}{\scriptsize CLEVR} & \rotatebox{90}{\scriptsize SST$2$} & \rotatebox{90}{\scriptsize ImageNet} & \rotatebox{90}{\scriptsize \bf Average} & \rotatebox{90}{\scriptsize \bf \# Wins.} \\
\toprule
CLIP~\cite{CLIP} & $55.7$ & $76.0$ & $46.9$ & $\mathbf{\underline{24.4}}$ & $50.7$ & $17.8$ & $4.8$ & $31.5$ & $53.7$ & $78.4$ & $31.8$ & $26.8$ & $37.6$ & $89.0$ & $22.7$ & $36.9$ & $\mathbf{\underline{24.1}}$ & $6.8$ & $20.0$ & $49.1$ & $40.3$ & $39.3$ & $2$ \\
SLIP~\cite{SLIP} & $56.7$ & $73.4$ & $43.2$ & $22.6$ & $51.6$ & $17.7$ & $4.9$ & $32.4$ & $52.5$ & $79.1$ & $33.3$ & $\mathbf{\underline{29.4}}$ & $33.5$ & $89.5$ & $17.8$ & $36.2$ & $17.8$ & $6.8$ & $\mathbf{\underline{23.4}}$ & $49.7$ & $41.6$ & $38.7$ & $2$ \\
\maeclip{} & $57.8$ & $78.2$ & $52.4$ & $23.9$ & $51.6$ & $18.1$ & $4.6$ & $31.5$ & $55.8$ & $78.4$ & $32.0$ & $27.6$ & $32.7$ & $89.8$ & $27.0$ & $39.4$ & $22.9$ & $7.2$ & $14.7$ & $49.3$ & $42.3$ & $39.9$ & $0$ \\
\midrule
\ourmethod & $\mathbf{\underline{58.9}}$ & $\mathbf{\underline{86.2}}$ & $\mathbf{\underline{55.1}}$ & $23.4$ & $\mathbf{\underline{51.8}}$ & $\mathbf{\underline{19.5}}$ & $\mathbf{\underline{5.9}}$ & $\mathbf{\underline{32.8}}$ & $\mathbf{\underline{59.2}}$ & $\mathbf{\underline{80.7}}$ & $\mathbf{\underline{33.5}}$ & $22.6$ & $\mathbf{\underline{40.1}}$ & $\mathbf{\underline{93.2}}$ & $\mathbf{\underline{28.8}}$ & $\mathbf{\underline{40.2}}$ & $19.1$ & $\mathbf{\underline{7.8}}$ & $16.8$ & $\mathbf{\underline{50.0}}$ & $\mathbf{\underline{45.0}}$ & $\mathbf{\underline{42.3}}$ & $\mathbf{\underline{17}}$ \\

\bottomrule
\end{tabular}}
\vspace{-0.3cm}
\caption{\textbf{Zero-shot classification on $\mathbf{21}$ datasets.} All models are trained with ViT-B/$16$ encoder for $25$ epochs on L-$20$M.}
\label{tab: zero_shot_classification}
\vspace{-0.3cm}
\end{table*}

\begin{table*}[tbp]
\centering
\tablestyle{3pt}{1.1}
\setlength{\tabcolsep}{2.6mm}{
\begin{tabular}{l|cccccccccccc}
\multirow{4}{*}{Method} & \multicolumn{6}{c}{COCO} & \multicolumn{6}{c}{Flickr$30$K} \\
\cmidrule(lr){2-7}\cmidrule(lr){8-13}
& \multicolumn{3}{c}{I$\rightarrow$T} & \multicolumn{3}{c}{T$\rightarrow$I} & \multicolumn{3}{c}{I$\rightarrow$T} & \multicolumn{3}{c}{T$\rightarrow$I} \\
\cmidrule(lr){2-4}\cmidrule(lr){5-7}\cmidrule(lr){8-10}\cmidrule(lr){11-13}
& R@$1$ & R@$5$ & R@$10$ & R@$1$ & R@$5$ & R@$10$ & R@$1$ & R@$5$ & R@$10$ & R@$1$ & R@$5$ & R@$10$ \\
\toprule
CLIP~\cite{CLIP} & $41.82$ & $69.50$ & $79.34$ & $30.54$ & $57.10$ & $69.30$ & $28.13$ & $50.77$  & $61.12$ & $20.61$ & $40.77$ & $50.49$ \\
SLIP~\cite{SLIP} & $44.54$ & $72.20$ & $82.10$ & $33.26$ & $59.66$ & $71.14$ & $31.20$ & $55.81$  & $66.02$ & $24.05$ & $45.58$ & $55.90$ \\
\maeclip{} & $42.72$ & $70.66$ & $80.36$ & $31.40$ & $57.50$ & $69.68$ & $28.62$ & $52.25$  & $62.64$ & $22.64$ & $43.67$ & $53.97$\\
\midrule
\ourmethod & $\textbf{\underline{45.06}}$ & $\textbf{\underline{73.38}}$ & $\textbf{\underline{83.36}}$ & $\textbf{\underline{34.86}}$ & $\textbf{\underline{61.36}}$ & $\textbf{\underline{72.78}}$ & $\textbf{\underline{32.21}}$ & $\textbf{\underline{56.39}}$  & $\textbf{\underline{66.67}}$ & $\textbf{\underline{25.48}}$ & $\textbf{\underline{47.55}}$ & $\textbf{\underline{57.94}}$ \\
\bottomrule
\end{tabular}}
\vspace{-0.2cm}
\caption{\textbf{Zero-shot image-text retrieval.} I$\rightarrow$T and T$\rightarrow$I indicate image-to-text retrieval and text-to-image retrieval.}
\vspace{-0.5cm}
\label{tab: zero_shot_retrieval}
\end{table*}

\begin{table}[!tbp]
\centering
\tablestyle{3pt}{1.1}
\setlength{\tabcolsep}{1.4mm}{
\begin{tabular}{l|ccccc|c}
Method & \rotatebox{90}{\footnotesize IN-A} &  \rotatebox{90}{\footnotesize IN-R} & \rotatebox{90}{\footnotesize IN-Ske} & \rotatebox{90}{\footnotesize IN-V$2$} & \rotatebox{90}{\footnotesize ObjNet} & \rotatebox{90}{\footnotesize \bf Average} \\
\toprule
CLIP~\cite{CLIP} & $9.3$ & $51.2$& $28.1$& $39.8$& $17.7$& $32.3$ \\
SLIP~\cite{SLIP} & $10.5$ & $49.8$& $26.7$& $41.3$& $20.4$& $33.1_{\uparrow0.8}$ \\
\maeclip{} & $11.6$ & $53.9$& $31.1$& $41.6$& $19.4$& $34.4_{\uparrow2.1}$ \\
\midrule
\ourmethod & $\mathbf{\underline{12.1}}$ & $\mathbf{\underline{55.7}}$& $\mathbf{\underline{31.4}}$& $\mathbf{\underline{43.3}}$& $\mathbf{\underline{21.0}}$& $\mathbf{\underline{35.7_{\uparrow3.4}}}$ \\
\bottomrule
\end{tabular}}
\vspace{-0.2cm}
\caption{\textbf{Zero-shot out-of-distribution classification.} We report the Top-$1$ classification accuracy ($\%$) for reference.}
\label{tab: zero_shot_ood}
\vspace{-0.5cm}
\end{table}

\noindent\textbf{Classification.}
We evaluate the zero-shot classification over $21$ benchmarks including ImageNet-$1$K~\cite{ImageNet}.
Detail of each datasets are listed in the appendix and the evaluate recipes (\eg, prompt engineering) strictly follow \cite{SLIP}.
Results are shown in Table~\ref{tab: zero_shot_classification}.
Specifically, compared to CLIP, \maeclip{} only achieves $+0.6\%$ average improvements, while our \ourmethod{} shows $+2.0\%$ gains.
This hints the masked image modeling objective in the na\"ive combination has little help to image-text alignment, while ours alleviate this issue by bind two objectives in a unified landscape.
On ImageNet-$1$K, our method achieves $45.0\%$ accuracy, $+4.7\%/+3.4\%/+2.7\%$ higher than CLIP, SLIP and \maeclip{}, respectively.
Among $21$ benchmarks, our method outperforms others over $17$ datasets, frequently with a significant margin.

\noindent\textbf{Image-Text Retrieval.}
We study image-text retrieval on $2$ benchmarks: COCO~\cite{COCO} and Flickr$30$K~\cite{Flickr30K}.
For both benchmarks, we use the original captions (w/o prompt) and $224\times224$ resized images for retrieval.
Different from zero-shot classification, retrieval requires the models to have the ability to capture fine-grained contextualized information in both images and texts.
As the results in Table~\ref{tab: zero_shot_retrieval} shown, our model achieves the best among all counterparts.
In particular, when compared to CLIP, our \ourmethod{} shows more significant improvements than \maeclip{}.

\noindent\textbf{Robustness to Distribution Shift.}
Following CLIP~\cite{CLIP}, we also conduct zero-shot experiments on $5$ out-of-distribution datasets: ImageNet Adversarial~\cite{ImageNet-Adversarial}, ImageNet Rendition~\cite{ImageNet-Rendition}, ImageNetV2~\cite{ImageNetV2}, ImageNet Sketch~\cite{ImageNet-Sketch}, and ObjectNet~\cite{ObjectNet}.
As shown in Table~\ref{tab: zero_shot_ood}, our method outperforms others on all five benchmarks.
Specifically, \ourmethod{} obtains significant improvements, $+3.4\%$, $+2.6\%$, $+1.3\%$ better than CLIP, SLIP, and \maeclip{} respectively.

\section{Ablation Study}

In this section, we ablate the designs of \ourmethod{}. All experiments are conducted with ViT/B-$16$ vision encoder and trained on L-$10$M for $25$ epochs.
We report the classification accuracy ($\%$) under zero-shot (ZS.), linear probing (Lin.) and end-to-end fine-tuning (FT.) on ImageNet-$1$K.

\subsection{Comparisons with Two-stage Methods}
As discussed above, another way to leverage both masked image modeling with image-text contrastive learning is to build a two-stage framework and learn two objectives step by step.
In this section, we compare our \ourmethod{} with several two-stage methods:

\noindent\textbf{MIM$\rightarrow$LiT} indicates firstly pre-train with masked image modeling only, then follow \cite{LiT} to perform locked-image text tuning on image-text pairs. Specifically, we start the second stage by fine-tuning pre-trained MAE~\cite{MAE}.

\noindent\textbf{MIM$\rightarrow$CLIP} denotes fully fine-tune pre-trained MAE on image-text pairs in the second stage. In this way, pre-trained model also inherit properties from both objectives.

\noindent\textbf{CLIP$\rightarrow$MIM} stands for using pre-trained CLIP as a guidance for masked image modeling in the second stage. This paradigm has been studied in recent research such as \cite{MVP, MILAN, BEiT2}.

The comparison results are shown in Table~\ref{tab: two_stage_comparison}.
As the vision encoder is fully frozen during the second stage in MIM$\rightarrow$LiT, its performance on downstream tasks remains unchanged except for zero-shot.
MIM$\rightarrow$CLIP slightly outperforms MAE and CLIP.
CLIP$\rightarrow$MIM exhibits more improvements upon two base methods, but lose the ability on zero-shot classification.
Our method rivals all counterparts with a more concise training pipeline.

\subsection{Comparisons on Reconstruction Space}
The core philosophy of our design is to perform masked reconstruction in language semantic space.
We ablate the effectiveness of our design by comparing to two other alternatives: raw pixel space and high-level vision space.
Reconstruction in raw pixel space denotes the aforementioned \maeclip{} which tries to reconstruct raw pixels directly.
For high-level vision space, we replace the language feature $z^{T}$ in Eq.~\eqref{equation: prob_distribution} to learnable weights with other components unchanged. In other words, similar to design in \cite{iBOT, MSN, DINO}, we map patch features to a probabilistic distribution on a group of learnable weights.
As results in Table~\ref{tab: ablation_on_reconstruction_target} shown, our \ourmethod{} shows better performance on all three metrics. We notice that compared to reconstruct raw pixels, the high-level space reconstruction does not shown prominent advantages as in \cite{DINO, iBOT, MSN}.
We guess this is due to the absence of multi-crop augmentation and exponential moving average (EMA)  strategy which play important roles inside these methods but bring significant adverse impact on training efficiency.

\begin{table}
\centering
\tablestyle{3pt}{1.1}
\setlength{\tabcolsep}{2mm}{
\begin{tabular}{l|ccc}
Method & ZS. & Lin. & FT. \\
\toprule
MAE~\cite{MAE} & $-$ & $43.4$ & $81.5$ \\
CLIP~\cite{CLIP} & $32.1$ & $64.1$ & $82.0$ \\
\midrule
MIM$\rightarrow$LiT~\cite{LiT} & $13.2$ & $43.4$ & $81.5$ \\
MIM$\rightarrow$CLIP & $34.4$ & $64.8$ & $82.2$ \\
CLIP$\rightarrow$MIM~\cite{MVP, MILAN, BEiT2} & $-$ & $66.2$ & $82.4$ \\
\midrule
\ourmethod{} (E$2$E) & $\mathbf{\underline{37.5}}$ & $\mathbf{\underline{68.5}}$ & $\mathbf{\underline{82.7}}$ \\
\bottomrule
\end{tabular}}
\vspace{-0.2cm}
\caption{\textbf{Comparisons with two-stage methods.} All methods are trained with ViT-B/$16$ on L-$10$M for $25$ epochs. Our approach rivals all two-stage methods in terms of zero-shot (ZS.), linear probing (Lin.) and end-to-end fine-tuning (FT.)}
\label{tab: two_stage_comparison}
\vspace{-0.3cm}
\end{table}

\begin{table}
\centering
\tablestyle{3pt}{1.1}
\setlength{\tabcolsep}{2mm}{
\begin{tabular}{l|ccc}
Reconstruction Space & ZS. & Lin. & FT. \\
\toprule
Raw Pixel Space (\maeclip{}) & $34.2$ & $61.9$ & $82.2$ \\
High-level Vision Space \cite{iBOT, SdAE} & $34.8$ & $67.7$ & $82.4$ \\
\midrule
Language Semantic Space (\ourmethod{}) & $\mathbf{\underline{37.5}}$ & $\mathbf{\underline{68.5}}$ & $\mathbf{\underline{82.7}}$ \\
\bottomrule
\end{tabular}}
\vspace{-0.2cm}
\caption{\textbf{Comparisons on reconstruction space.} Raw pixel space denotes aforementioned \maeclip{} which directly reconstruct RGB values. High-level vision space indicates to replace the language feature $z^{T}$ in Eq.~\eqref{equation: prob_distribution} to a group of learnable embeddings similar to \cite{iBOT, DINO, MSN}.  Language semantic space stands for ours. All methods are trained with ViT-B/$16$ for 25 epochs on L-$10$M.}
\label{tab: ablation_on_reconstruction_target}
\vspace{-0.5cm}
\end{table}

\subsection{Hyper-parameters Analysis}
\noindent\textbf{Mask Ratio.}
We compare three different mask ratios in Table~\ref{tab: ablation_mask_ratio}.
As results shown, randomly masking at a ratio of $75\%$ leads to generally decent performance.
Lower mask ratio ($60\%$) leads to $+0.8\%$ better linear probing accuracy but impairs zero-shot performance ($-1.6\%$).
Higher mask ratio ($90\%$) acquires similar zero-shot accuracy but results in apparent decrease of $-0.9\%$ and $-0.4\%$ in terms of linear probing and fine-tuning.
We guess this is due to lower mask ratio brings more prior information but less supervision in reconstruction, while the higher one exact converses.

\noindent\textbf{Number of vision decoder blocks.}
Performance \wrt \ number of decoder blocks are shown in Table~\ref{tab: ablation_decoder_blocks}.
Different decoder blocks numbers exhibit less to none differences in terms of zero-shot.
As to linear probing, we detect the same trend as \cite{MAE}: More vision decoder blocks leads to better linear probing accuracy.
For fine-tuning, we observe almost the same accuracy but more decoder blocks seems to be slightly worse.
Besides the numerical differences, when take the extra costs bring by more decoder blocks into consideration, we choose to build our \ourmethod{} with one vision decoder blocks only, due to more decoder blocks introduces un-negligible training overheads.
For instance, increasing the number of decoder blocks from $1$ to $4$ brings $0.5\times$ more training costs.

\noindent\textbf{Loss coefficients.}
The results under different loss coefficients in Eq.~\eqref{equation: total_loss} are shown in Table~\ref{tab: ablation_loss_coefficients}.
We empirically observe better performance with $\lambda_{1}$:$\lambda_{2}$ setting to $2$:$1$.

\begin{table}[tbp]
\tablestyle{3pt}{1.1}
\begin{subtable}[t]{1.0\linewidth}
\centering
\setlength{\tabcolsep}{4.0mm}{
\begin{tabular}[t]{cccc}
Mask Ratio & ZS. & Lin. & FT. \\
\toprule
$60\%$ & $35.9$ & $69.3$ & $82.7$ \\
\rowcolor{defaultcolor} $75\%$ & $37.5$ & $68.5$ & $82.7$ \\
$90\%$ & $37.4$ & $67.6$ & $82.3$ \\
\bottomrule
\end{tabular}}
\captionsetup{width=0.9\linewidth}
\caption{\textbf{Ablations on mask ratio.} $75\%$ generally leads to a good result except for linear probing.}
\label{tab: ablation_mask_ratio}
\end{subtable}
\begin{subtable}[t]{1.0\linewidth}
\centering
\setlength{\tabcolsep}{3.0mm}{
\begin{tabular}[t]{ccccc}
Nums. &  ZS. & Lin. & FT. & Rel.GHs. \\
\toprule
\rowcolor{defaultcolor} $1$ & $37.5$ & $68.5$ & $82.7$ & $1.0\times$ \\
$2$ & $37.8$ & $68.9$ & $82.6$ & $1.3\times$ \\
$4$ & $37.3$ & $69.6$ & $82.5$ & $1.5\times$ \\
\bottomrule
\end{tabular}}
\captionsetup{width=0.9\linewidth}
\caption{\textbf{Numbers of vision decoder blocks.} Rel.GHs. is short for relative GPU hours.}
\label{tab: ablation_decoder_blocks}
\end{subtable}
\begin{subtable}[t]{1.0\linewidth}
\centering
\setlength{\tabcolsep}{4.5mm}{
\begin{tabular}[t]{cccc}
$\lambda_1$:$\lambda_2$ & ZS. & Lin. & FT. \\
\toprule
\rowcolor{defaultcolor} $2$:$1$ & $37.5$ & $68.5$ & $82.7$ \\
$1$:$1$ & $36.4$ & $68.1$ & $82.3$ \\
$1$:$2$ & $35.9$ & $68.2$ & $82.2$ \\
\bottomrule
\end{tabular}}
\captionsetup{width=0.9\linewidth}
\caption{\textbf{Loss coefficients} of $\mathcal{L}_{\mathrm{Contra}}$ and $\mathcal{L}_{\mathrm{Recon}}$.}
\label{tab: ablation_loss_coefficients}
\end{subtable}
\vspace{-0.3cm}
\caption{Ablations on the hyper-parameters of our approach. We report zero-shot (ZS.), linear probing(Lin.) and end-to-end fine-tuning (FT.) accuracy ($\%$) on ImageNet-$1$K. Our default setups are high-lighted in \colorbox{defaultcolor}{gray}.}
\label{tab: ablation_studies}
\vspace{-0.5cm}
\end{table}

\section{Conclusion}
In this work, we present a unified vision representation learner \ourmethod{} that subsumes masked image modeling with natural language supervision.
Instead of simply combining both paradigms with limited connection, we present a novel design to perform MIM in the language semantic space.
Text features from language encoder serves as basic prototypes and probabilistic distribution of masked patches explicitly .
Both objectives complement each other, leading to high synergy and mutual benefits.
Extensive experimental results on downstream tasks showcase our method's advanced transferability and out-of-the-box capacity.
We also observe excellent properties emerge from our design, especially in the low-shot regime.
We hope our work can provide a new perspective on how to utilize language supervision with masked image modeling.
In the future, we will explore further scale-up of our approach in terms of both model size and data size.

{\small
\bibliographystyle{ieee_fullname}
\bibliography{egbib}
}

\clearpage
\appendix

\renewcommand{\thefigure}{A\arabic{figure}}
\setcounter{figure}{0}
\renewcommand{\thetable}{A\arabic{table}}
\setcounter{table}{0}

\section{Wall-clock Time Comparison}
\begin{table}[H]
\centering
\tablestyle{3pt}{1.1}
\setlength{\tabcolsep}{2mm}{
\begin{tabular}{lcc|ccc}
Method & Dataset & PT Epo. & ZS. & FT. & Rel.GHs. \\
\toprule
MAE~\cite{MAE} & \multirow{4}{*}{LAION-$20$M} & \multirow{4}{*}{$25$} & - & $82.1$ & $1.0\times$ \\
CLIP~\cite{CLIP} & & & $40.3$ & $82.7$ & $1.5\times$ \\
SLIP~\cite{SLIP} & & & $41.6$ & $82.6$ & $3.7\times$ \\
\maeclip{} & & & $42.3$ & $82.9$ & $1.8\times$ \\
\midrule
\ourmethod{} & LAION-$20$M & $25$ & $\mathbf{\underline{45.0}}$ & $\mathbf{\underline{83.3}}$ & $1.8\times$ \\
\bottomrule
\end{tabular}}
\vspace{-.5em}
\caption{\textbf{Wall-clock time comparison.} We report zero-shot (ZS.) and end-to-end fine-tuning (FT.) accuracy on ImageNet-$1$K for reference. Rel.GHs. denotes relative GPU hours. Compared to CLIP, our method only brings $20\%$ extra training time costs. Compared to \maeclip{}, our method exhibits better performance under the same training overhead.}
\label{tab: wall_clock_time_comparisons}
\vspace{-.5em}
\end{table}

\section{Data Scaling}
\begin{table}[H]
\centering
\tablestyle{3pt}{1.1}
\setlength{\tabcolsep}{2mm}{
\begin{tabular}{ccc|ccc}
\multirow{1}{*}{Method} & \multirow{1}{*}{Dataset} & \multirow{1}{*}{PT Epo.} & ZS. & Lin. & FT. \\
\toprule
\multirow{5}{*}{\ourmethod{}} & LAION-$10$M & \multirow{5}{*}{$25$} & $37.5$ & $68.5$ & $82.7$ \\
& YFCC-$15$Mv$2$~\cite{DeCLIP} & & $41.5$ & $70.2$ & $82.9$ \\
& LAION-$20$M & & $45.0$ & $71.5$ & $83.3$ \\
& LAION-$50$M & & $49.4$ & $71.9$ & $83.6$ \\
& LAION-$100$M & & $50.6$ & $72.2$ & $83.7$ \\
\bottomrule
\end{tabular}}
\vspace{-.5em}
\caption{\textbf{Scaling property of our \ourmethod{}.} All models are pre-trained with ViT-B/$16$~\cite{ViT} as vision encoder for $25$ epochs, and report zero-shot (ZS.), linear probing (Lin.) and fine-tuning (FT.) classification accuracy on ImageNet-$1$K. We observe contiguous gains when our approach meets more image-text pairs. Besides, we notice that increasing data from $50$M to $100$M shows relatively minor improvements, we speculate this is due to only scaling dataset instead of jointly scale-up dataset with model size. We leave this exploration in the future.}
\label{tab: data_scaling}
\vspace{-.5em}
\end{table}

\section{Implementation Details}

\subsection{Model Architecture Details}

\begin{table}[H]
\centering
\tablestyle{3pt}{1.1}
\setlength{\tabcolsep}{6.1mm}{
\begin{tabular}{l|l|l}
& Configuration & Value \\
\toprule
\multirow{6}{*}{Vision Encoder} & Patch Size & $16\times16$ \\
& Layers & $12$ \\
& Width & $768$ \\
& Heads & $12$ \\
& MLP Ratio & $4.0$ \\
& \# Parameters & $85.8$M \\
\midrule
\multirow{5}{*}{Vision Decoder} & Layers & $1$ \\
& Width & $768$ \\
& Heads & $12$ \\
& MLP Ratio & $4.0$ \\
& \# Parameters & $8.4$M \\
\midrule
\multirow{5}{*}{Language Encoder} & Layers & $12$ \\
& Width & $512$ \\
& Heads & $8$ \\
& MLP Ratio & $4.0$ \\
& \# Parameters & $37.8$M \\
\bottomrule
\end{tabular}}
\vspace{-.5em}
\caption{\textbf{Model architecture details.}}
\label{tab: model_architecture}
\vspace{-.5em}
\end{table}

\subsection{Pre-training}

\begin{table}[H]
\centering
\tablestyle{3pt}{1.1}
\setlength{\tabcolsep}{5.1mm}{
\begin{tabular}{l|l}
Configuration & Value \\
\toprule
Batch Size & $4096$ \\
Vocabulary Size & $49408$ \\
Training Epochs & $25$ \\
Optimizer & AdamW~\cite{Adam} \\
Learning Rate & $5$e-$4$ \\
Minimal Learning Rate & $1$e-$5$ \\
Weight Decay & $0.5$ \\
Adam $\beta_1$ & $0.9$ \\
Adam $\beta_2$ & $0.98$ \\
Warmup Epochs & $1$ \\
Learning Rate Schedule & Cosine \\
Augmentation & RandomResizedCrop($0.5$, $1.0$) \\
Mask Ratio & $0.75$ \\
\bottomrule
\end{tabular}}
\vspace{-.5em}
\caption{\textbf{Pre-training settings.}}
\label{tab: pretraining_settings}
\vspace{-.5em}
\end{table}

\subsection{ImageNet-1K Fine-tuning}
\begin{table}[H]
\centering
\tablestyle{3pt}{1.1}
\setlength{\tabcolsep}{9.0mm}{
\begin{tabular}{l|l}
Configuration & Value \\
\toprule
Batch Size & $1024$ \\
Training Epochs & $100$ \\
Optimizer & AdamW~\cite{Adam} \\
Learning Rate & $4$e-$4$ \\
Weight Decay & $0.05$ \\
Adam $\beta_1$ & $0.9$ \\
Adam $\beta_2$ & $0.999$ \\
Warmup Epochs & $5$ \\
Learning Rate Schedule & Cosine \\
Layer-wise LR Decay & $0.65$ \\
Augmentation & RandAug($9$, $0.5$)~\cite{RandAug} \\
Label Smoothing & $0.1$ \\
Mixup & $0.8$ \\
CutMix & $1.0$ \\
Drop Path & $0.1$ \\
\bottomrule
\end{tabular}}
\vspace{-.5em}
\caption{\textbf{ImageNet-1K fine-tuning settings.}}
\label{tab: imagenet_finetuning_setups}
\vspace{-.5em}
\end{table}
\subsection{Semantic Segmentation Fine-tuning}
\begin{table}[H]
\centering
\tablestyle{3pt}{1.1}
\setlength{\tabcolsep}{10.7mm}{
\begin{tabular}{l|l}
Configuration & Value \\
\toprule
Batch Size & $16$ \\
Training Iters & $160$K \\
Optimizer & AdamW~\cite{Adam} \\
Learning Rate & $1$e-$4$ \\
Weight Decay & $0.05$ \\
Adam $\beta_1$ & $0.9$ \\
Adam $\beta_2$ & $0.999$ \\
Warmup Iters & $1500$ \\
Learning Rate Schedule & Poly \\
Layer-wise LR Decay & $0.65$ \\
Image Size & $512\times512$ \\
\bottomrule
\end{tabular}}
\vspace{-.5em}
\caption{\textbf{ADE20K fine-tuning settings.}}
\label{tab: ade20k_setups}
\vspace{-.5em}
\end{table}
\subsection{Detection Fine-tuning}
\begin{table}[H]
\centering
\tablestyle{3pt}{1.1}
\setlength{\tabcolsep}{5.6mm}{
\begin{tabular}{l|cc}
Configuration & COCO~\cite{COCO} & LVIS~\cite{LVIS} \\
\toprule
Batch Size & \multicolumn{2}{c}{$16$} \\
Training Epochs & \multicolumn{2}{c}{$25$} \\
Optimizer & \multicolumn{2}{c}{AdamW~\cite{Adam}} \\
Learning Rate & $1$e-$4$ & $2$e-$4$ \\
Weight Decay & \multicolumn{2}{c}{$0.1$} \\
Adam $\beta_1$ & \multicolumn{2}{c}{$0.9$} \\
Adam $\beta_2$ & \multicolumn{2}{c}{$0.999$} \\
Warmup Iters & \multicolumn{2}{c}{$250$} \\
Learning Rate Schedule & \multicolumn{2}{c}{Cosine} \\
Layer-wise LR Decay & \multicolumn{2}{c}{$0.7$} \\
Drop Path & \multicolumn{2}{c}{$0.1$} \\
Image Size & \multicolumn{2}{c}{$1024\times1024$} \\
Augmentation & \multicolumn{2}{c}{LSJ($0.1, 2.0$)~\cite{CopyPaste}} \\
\bottomrule
\end{tabular}}
\vspace{-.5em}
\caption{\textbf{COCO and LVIS fine-tuning settings.}}
\label{tab: coco_lvis_setups}
\vspace{-.5em}
\end{table}

\section{Pre-training Pseudo Code}
\begin{algorithm}[H]
\caption{\ourmethod{} pre-training pseudo-code in PyTorch style.}
\label{algo: pytorch}
\definecolor{codegreen}{rgb}{0,0.5,0}
\definecolor{codeblue}{rgb}{0.25,0.5,0.5}
\definecolor{codegray}{rgb}{0.6,0.6,0.6}
\lstset{
  backgroundcolor=\color{white},
  basicstyle=\fontsize{7.5pt}{8.5pt}\fontfamily{lmtt}\selectfont,
  columns=fullflexible,
  breaklines=true,
  captionpos=b,
  commentstyle=\fontsize{8pt}{9pt}\color{codegray},
  keywordstyle=\fontsize{8pt}{9pt}\color{codegreen},
  stringstyle=\fontsize{8pt}{9pt}\color{codeblue},
  frame=tb,
  otherkeywords = {self},
}
\begin{lstlisting}[language=python]
# xi, xt: input images and texts
# v_enc, v_dec: vision encoder, vision decoder
# l_enc: language encoder
# v_proj, l_proj: vision and language projector
# sigma, tau1, tau2: temperatures
# lambda_1, lambda_2: loss coefficients
# B, N, D: batch size, patch numbers, feature dimension

def forward(xi, xt):

    #random mask input images
    masked_xi = random_mask(xi, mask_ratio=0.75)
	
    zi = v_enc(xi) #[B, N, D]
    masked_zi = v_enc(masked_xi)
    gi = v_dec(masked_zi) #[B, N, D]
    zt = l_enc(xt) #[B, D]
	
    return forward_loss(zi, gi, zt)

def forward_loss(zi, gi, zt):
    
    #vision language contrastive
    ei = norm(v_proj(zi.mean(dim=1))) #[B, D]
    et = norm(l_proj(zt)) #[B, D]
    
    label = range(B)
    logit = ei @ et.T / sigma #[B, B]
    
    i2t = cross_entropy(logit, label)
    t2i = cross_entropy(logit.T, label)
    l_contra = (i2t + t2i) / 2.
    
    #masked visual reconstruction
    zi = norm(v_proj(zi)) #[B, N, D]
    gi = norm(v_proj(gi)) #[B, N, D]
    
    logit_p = (gi @ zt.T / tau1).softmax(-1) #[B, N, B]
    logit_t = (zi @ zt.T / tau2).softmax(-1) #[B, N, B]
    
    #reconstruction in language semantic space
    l_recon = kl_divergence(logit_p, logit_t)
    
    return lambda_1 * l_contra + lambda_2 * l_recon
	
\end{lstlisting}
\end{algorithm}

\end{document}